# Automated Genre-Aware Article Scoring and Feedback Using Large Language Models

Chihang Wang
New York University
New York, USA

Yuxin Dong
Wake Forest University
Winston-Salem, USA

Zhenhong Zhang
George Washington University
Washington, USA

Ruotong Wang
University at Albany, State University of New York
Albany, USA

Shuo Wang
Purdue University, Indianpolis
Indianapolis, USA

Jiajing Chen*
New York University
New York, USA

*Abstract*—This paper focuses on the development of an advanced intelligent article scoring system that not only assesses the overall quality of written work but also offers detailed feature-based scoring tailored to various article genres. By integrating the pre-trained BERT model with the large language model Chat-GPT, the system gains a deep understanding of both the content and structure of the text, enabling it to provide a thorough evaluation along with targeted suggestions for improvement. Experimental results demonstrate that this system outperforms traditional scoring methods across multiple public datasets, particularly in feature-based assessments, offering a more accurate reflection of the quality of different article types. Moreover, the system generates personalized feedback to assist users in enhancing their writing skills, underscoring the potential and practical value of automated scoring technologies in educational contexts.

*Keywords-Automated article scoring, Large language models, Feature scoring, Educational technology*

## I. Introduction

In the current context of globalization, English, as the official language of the world, plays an irreplaceable and key role. Writing, as the core link of a comprehensive assessment of students' English proficiency, can deeply demonstrate students' proficiency and mastery of English vocabulary, grammar, etc. [1]. In the test paper, English essays account for a relatively high proportion, which makes the grading work more arduous [2]. At the same time, the traditional manual grading process is often greatly affected by the subjective emotions of the grader [3]. In recent years, with the rapid rise of online education, automated article grading, as part of automated evaluation, must keep up with the times and aim to reduce the heavy burden of English teachers in grading [4]. In order to meet the challenge, the automated article grading algorithm uses advanced technologies such as natural language translation and semantic understanding and matching, becoming a forward-looking and practical application in the field of natural language processing (NLP) [5].

The development of automated article grading algorithms is closely dependent on the rapid innovation of natural language technology, which has been widely applied in fields such as disease prediction [6-8] and risk management[9-11], achieving notable results. Natural language technology has always focused on enabling computers to understand, interpret and generate natural language [12]. Some studies [13,14] have attempted to represent articles from a multi-task perspective and proposed a method based on a bidirectional recurrent neural network. The educational application of language models in automated article scoring is a striking research direction in the field of educational technology. With the advancement of education digitization, more and more students and educators use online learning platforms and educational applications for learning and teaching. Automated article scoring technology plays an important role in this context, providing efficient, rapid, and personalized evaluation, and contributing to the optimal use of educational resources [15].

This paper proposes an automated scoring feedback algorithm for articles based on different genres. It mainly solves the problem of ignoring the genre features of articles. Not only does it score the article as a whole, but it also achieves a more objective and comprehensive article evaluation by introducing feature scoring of articles of different genres. This genre-aware scoring mechanism successfully captures the unique features of different types of articles, thereby improving the accuracy of scoring. Further introducing a large language model for automated article scoring tasks, the algorithm not only outputs the overall quality score of the article but also provides students with specific writing guidance to help them understand and improve their performance on different features. The performance of the algorithm has been verified on the public dataset ASAP++.

## II. METHOD

One of the problems with current automatic scoring methods is that they fail to fully consider the differences in the characteristics of English articles in different genres. Taking narrative English essays as an example, the focus of evaluation may be mainly on the strength of the narrative, while for explanatory English essays, more attention should be paid to the rationality of the organizational structure. Most current scoring methods ignore capturing the unique writing characteristics of different genres, resulting in a lack of accuracy and pertinence in the evaluation results.

Another problem is that most automated scoring algorithms for articles do not combine the scoring of articles with the corresponding evaluation feedback. This relatively one-sided evaluation method prevents students from getting specific guidance and improvement directions, especially when facing their weaknesses in certain article genres or characteristics. The disconnection between scoring results and evaluation feedback hinders students from fully understanding the evaluation results and limits their targeted improvement of individual writing skills. In order to solve the above problems, this paper proposes to organically combine scoring with feedback, which is an important step to improve the accuracy and practicality of automated scoring algorithms for articles. Therefore, an automated scoring algorithm for articles based on the BERT model is introduced [16], which pays special attention to the characteristics of the article as a whole and articles of different genres. Its overall architecture is shown in Figure 1

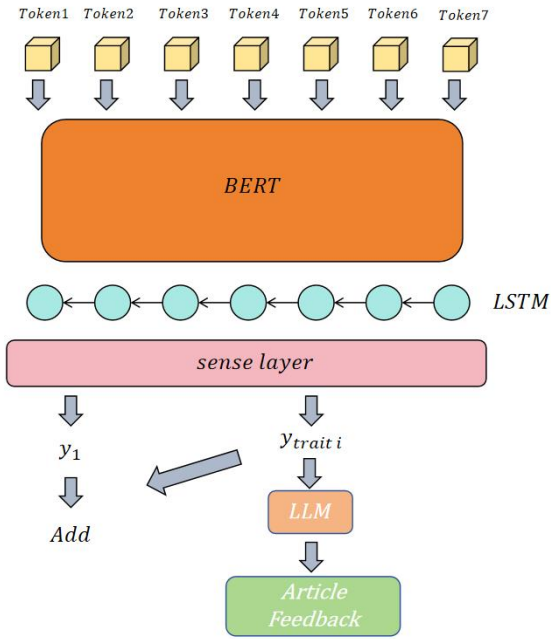

Figure 1 Overall architecture

The specific execution process of this module is as follows: First, select the 4,000 most frequently used words as the vocabulary, and use BERT's Word piece tokenizer to tokenize the input text $T$ of length $n$. The tokenization process is to separate the word itself from the prefix and suffix [17]. Each input text is followed by a text score. While tokenizing, add a separator [CLS] to the front of each text sequence $T$, and a separator [SEP] between each sentence. Under the maximum text length limit of the BERT model, add a filler [PAD] to texts that are insufficient in length, and truncate texts that exceed the length to obtain a sequence $T1$. The calculation formula is:

$$T1 = \begin{cases} [CLS] + T_L + [SEP] & n > L \\ [CLS] + T + [SEP] & n = L \\ [CLS] + T + [PAD]*(L-l) + [SEP] & n < L \end{cases}$$

Next, the obtained sequence $T1$ is input into word embedding, segment embedding, and position embedding. Word embedding converts the words in T1 into vectors of fixed dimensions, denoted as:

$$T_t = f_{TE}(T_1)$$

Segment embedding is to assign different values to words to distinguish different sentences, which can be written as:

$$T_s = f_{SE}(T_1)$$

Position embedding is to encode the words at the same position in each sentence in the same position [18], thereby marking the order information of the word sequence in the text, recorded as:

$$T_p = f_{PE}(T_1)$$

The final input vector T2 of the pre-trained model is the sum of the results of the above three embedding layers, denoted as:

$$T_2 = f_{TE}(T_1) + f_{SE}(T_1) + f_{PE}(T_1)$$

The semantic representation learned by the BERT model through pre-training is very rich, which can better capture the context, contextual relations and semantic information in the article. At the same time, it has strong adaptability to articles of various types and topics, making this method superior when processing articles in different fields. In the output of the BERT model, the output of the [CLS] tag contains the semantic representation of the article as a whole. The bidirectional encoding mechanism of the BERT model enables it to consider all words in the context [19-20], not just the information within the local window. This helps to better understand the overall semantics of the article, so by inputting the output of [CLS] into the fully connected layer, the overall features of the article can be understood and scored to the greatest extent. This connection method simplifies the entire model structure and directly uses BERT as a feature extractor, making the model easier to understand, train, and deploy [21-22].

The main structure of the article feedback output module is the chat-GPT model. The specific model structure is shown in Figure 2.

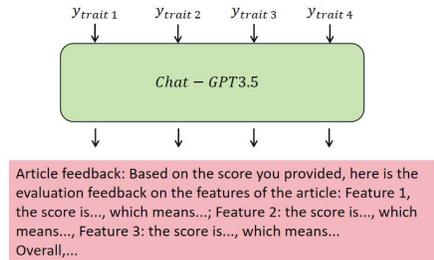

Figure 2 Specific model structure

The large language model will conduct a comprehensive analysis of the article content and feature scores. The large language model can capture multiple aspects of information such as the article's language structure, sentiment, logical organization, etc., while paying attention to the context related to the feature score. Ultimately, the large language model will generate detailed feedback for different feature scores. This feedback mechanism based on the large language model enables the module to understand the article at a deeper level and provide personalized suggestions for each feature score. It not only helps to understand the quality of the article, but also provides authors with specific directions for improvement, making the module more instructive and intelligent in the task of automated article scoring, and providing feedback and evaluation of the article from both quantitative and qualitative dimensions.

### III. EXPERIMENT

#### A. Datasets

The dataset used by the algorithm is the open-source dataset Automated Student Assessment Prize (ASAP) [23]. The dataset contains eight article collections. The articles in each collection are written by middle school students based on a prompt. The dataset contains nearly 13,000 articles, including three article genres: argumentative essays, question-answering essays, and narrative essays. The original dataset has scores for the entire article and scores for article features, but only the 7th and 8th collections provide different feature scores. In order to take into account the features of articles of different genres in the process of automated article scoring, this paper uses the dataset ASAP++ proposed by Mathias and Bhattacharyya, in which each article collection is supplemented with scores for about 4-6 different article features. The basic properties of the articles in the dataset are shown in Table 1. In order to further display the data, we provide a visual display of the data as shown in Figure 3.

Table 1 Dataset attribute table

| Article Collection | Number of words | Number of features |
|---|---|---|
| 1 | 350 | 5 |
| 2 | 350 | 5 |
| 3 | 100 | 4 |
| 4 | 100 | 4 |
| 5 | 125 | 4 |
| 6 | 150 | 4 |
| 7 | 300 | 4 |
| 8 | 600 | 6 |

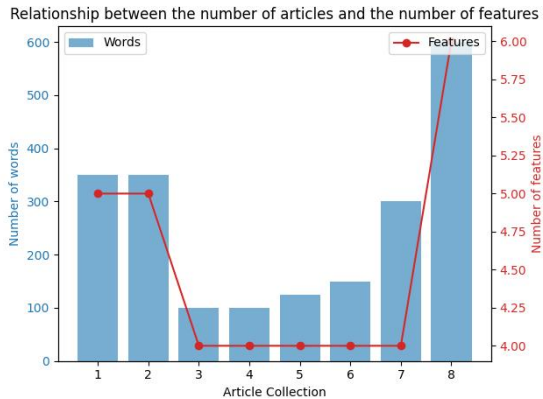

Figure 3 Relationship between the number of articles and the number of features

#### B. Experimental Results

The evaluation index used in this paper is the quadratic weighted Kappa value QWK. On the one hand, it can be compared with existing models more efficiently. On the other hand, the quadratic weighted Kappa value is credible when evaluating the results of the automated scoring algorithm of our article. The greater the error between the model prediction score and the actual score, the more severe the penalty, and the smaller the value of the QWK coefficient. The experimental results are shown in Table 2.

Table 2 Experimental results

| Number | Model | QKW |
|---|---|---|
| 1 | MHMLW | 0.732 |
| 2 | NFA | 0.741 |
| 3 | LC-A | 0.752 |
| 4 | SKIP-LSTM | 0.753 |
| 5 | CCXLNET | 0.761 |
| 6 | BERT-DOC-TOK-SEG | 0.762 |
| 7 | Tran-BERT-MS-ML-R | 0.793 |
| 8 | Ours | 0.803 |

Judging from the experimental results provided, our proposed model performed best among all tested models, reaching a QKW index of 0.803, which shows that compared to other models, our proposed model has better performance on the evaluated tasks. High accuracy or better performance. From MHMLW's 0.732 to NFA, LC-A, SKIP-LSTM, CCXLNET and BERT-DOC-TOK-SEG, we can observe a gradual improvement trend in performance, finally reaching 0.793 of Tran-BERT-MS-ML-R. Although each of these models has demonstrated its own advantages, in terms of final performance, our proposed model still achieves significant surpasses, which may be attributed to factors such as model design, training strategy, or data preprocessing methods used. These improvements are not only reflected in absolute values, but more importantly, they are improved by more than 0.01 compared to the second place Tran-BERT-MS-ML-R, showing that the model has stronger capabilities in handling specific tasks.

The ASAP++ dataset contains three different types of articles. Sets 1 and 2 are argumentative essays, sets 3 to 6 are answers to questions, and sets 7 and 8 are narrative essays. In order to verify the performance of the algorithm in different genres, three sets of articles 2, 3, and 8 of different article genres were randomly selected, and the QWK values of the three sets were analyzed. The three models with better QWK values were selected: Hierarchical LSTM-CNN-Attention, SKIPFLOW LSTM, and BERT-DOC-TOK-SEG. The reason why the best result Tran-BERT-MS-ML-R was not selected is that the structure of this model is the same as that of BERT-DOC-TOK-SEG, but there is a difference in the definition of the loss function.

Table 3 Model experimental results in the Laptop dataset

| Model | Collection 2 | Collection 3 | Collection 8 | Average |
|---|---|---|---|---|
| 3 | 0.683 | 0.692 | 0.732 | 0.702 |
| 4 | 0.687 | 0.695 | 0.754 | 0.712 |
| 6 | 0.691 | 0.699 | 0.776 | 0.722 |
| Ours | 0.701 | 0.703 | 0.804 | 0.736 |

In order to further demonstrate the superiority of our model, a bar chart of the corresponding experimental results is given here, as shown in Figure 4

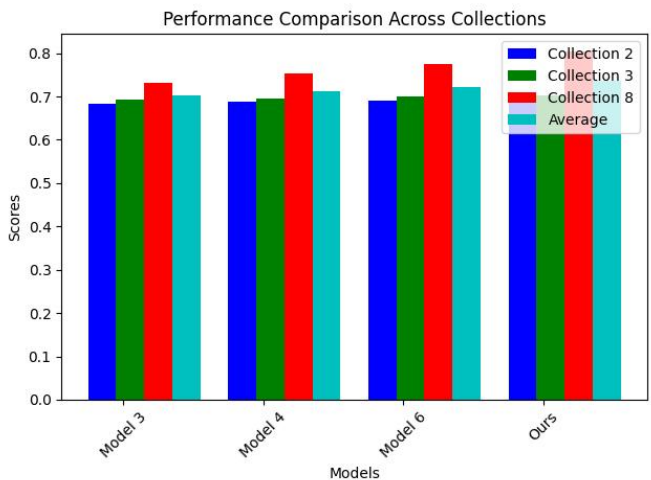

Figure 4 Experimental results

Judging from the experimental results provided, our model is significantly better than the other three models (Model 3, Model 4, Model 6) on all evaluation metrics. Especially on the Collection 8 data set, our model reached a score of 0.804, which is significantly improved compared to other models, which shows that our model has a stronger generalization ability and a more stable detection effect. The average score also reflects this trend, with our model achieving an average score of 0.736, approximately 1.4 to 3.4 percentage points higher than other models, demonstrating the consistent superiority of our model on a variety of data sets.

## IV. CONCLUSION

This paper presents an innovative automated article scoring and feedback system that significantly contributes to advancements in the broader field of artificial intelligence (AI). By integrating cutting-edge models such as BERT and Chat-GPT, the system enhances the traditional approach to article scoring through a sophisticated feature-based mechanism that adapts to the specific characteristics of different genres. This approach addresses a key limitation in prior models that treated articles as uniform, thereby improving both the accuracy and relevance of the scoring process. Experimental results, especially from the ASAP++ dataset, confirm that the model consistently outperforms existing systems across multiple dimensions, demonstrating AI's ability to analyze and understand complex textual data in ways that mimic human evaluators. Furthermore, the system's capacity to generate personalized feedback exemplifies the evolving role of AI in offering not just assessments but also actionable insights that help users enhance their writing. This represents a broader trend in AI where machines are not only performing analytical tasks but also facilitating learning and improvement through intelligent, context-aware interactions. By reducing the burden of manual grading and providing detailed, tailored feedback, this system highlights AI's potential to transform educational practices and other domains that rely on content evaluation. The work underscores how AI-driven tools are becoming increasingly sophisticated, offering practical and scalable solutions across various fields, and contributing to the continuous evolution of AI in understanding, assessing, and improving human communication. This development has far-reaching implications beyond education, indicating a growing reliance on AI for tasks requiring deep comprehension and nuanced judgment, ultimately pushing the boundaries of what AI can achieve in content analysis and feedback generation.